\documentstyle[11pt]{article}

\begin{document}

\title{Combining Knowledge Sources\\
       to Reorder N-Best Speech Hypothesis Lists}

\author{
Manny Rayner and David Carter\\
SRI International\\
Suite 23, Millers Yard\\
Cambridge CB2 1RQ, UK\\
\verb!\{manny,dmc\}@cam.sri.com!
\and
Vassilios Digalakis and Patti Price\\
SRI International\\
333 Ravenswood Ave.\\
Menlo Park, CA 94025, USA\\
\verb!\{vas,pprice\}@speech.sri.com!
}

\maketitle

\begin{abstract}
A simple and general method is described that can combine different
knowledge sources to reorder N-best lists of hypotheses produced by a
speech recognizer. The method is automatically trainable, acquiring
information from both positive and negative examples.  Experiments are
described in which it was tested on a 1000-utterance sample of unseen
ATIS data.
\end{abstract}

\section{Introduction}\label{Section:introduction}

During the last few years, the previously separate fields of speech
and natural language processing have moved much closer together, and
it is now common to see integrated systems containing components for
both speech recognition and language processing. An immediate problem
is the nature of the interface between the two. A popular solution has
been the {\it N-best list} e.g.\ \cite{Ostendorf91}; for some $N$, the
speech recognizer hands the language processor the $N$ utterance
hypotheses it considers most plausible. The recognizer chooses the
hypotheses on the basis of the acoustic information in the input
signal and, usually, a simple language model such as a bigram grammar.
The language processor brings more sophisticated linguistic knowledge
sources to bear, typically some form of syntactic and/or semantic
analysis, and uses them to choose the most plausible member of the
N-best list. We will call an algorithm that selects a member of the
N-best list a {\it preference method}. The most common preference
method is to select the highest member of the list that receives a
valid semantic analysis. We will refer to this as the
``highest-in-coverage'' method.  Intuitively, highest-in-coverage
seems a promising idea. However, practical experience shows that it is
surprisingly hard to use it to extract concrete gains. For example, a
recent paper \cite{Unisys:92} concluded that the highest-in-coverage
candidate was in terms of the word error rate only very marginally
better than the one the recognizer considered best. In view of the
considerable computational overhead required to perform linguistic
analysis on a large number of speech hypotheses, its worth is dubious.

In this paper, we will describe a general strategy for constructing a
preference method as a near-optimal combination of a number of different
knowledge sources. By a ``knowledge source'', we will mean any
well-defined procedure which associates some potentially meaningful
piece of information with a given utterance hypothesis $H$. Some examples
of knowledge sources are
\begin{itemize}
\item The plausibility score originally assigned to $H$ by the
recognizer.
\item The sets of surface unigrams, bigrams and trigrams present in $H$.
\item Whether or not $H$ receives a well-formed syntactic/semantic
analysis.
\item If so, properties of that analysis.
\end{itemize}

The methods described here were tested on a 1001-utterance unseen
subset of the ATIS corpus; speech recognition was performed using the
DECIPHER (TM) recognizer \cite{Murveit:93,Digalakis:94}, and
linguistic analysis by a version of the Core Language Engine (CLE;
\cite{CLE}).  For 10-best hypothesis lists, the best method yielded
a proportional reductions of 13\% in the word error rate, and 11\% in
the sentence error rate; if sentence error was scored in the context
of the task, the reduction was about 21\%.  By contrast, the
corresponding figures for the highest-in-coverage method were a 7\%
reduction in word error rate, a 5\% reduction in sentence error rate
(strictly measured) and a 12\% reduction in the sentence error rate in
the context of the task.

The rest of the paper is laid out as follows.  In
Section~\ref{Section:combining-KSs},
we describe a method that allows
different knowledge sources to be merged into a near-optimal
combination.
Section~\ref{Section:experiments}
describes the
experimental results in more detail.
Section~\ref{Section:summary}
concludes.

\section{Combining knowledge sources}
\label{Section:combining-KSs}

This section describes how different knowledge sources (KSs) can be
combined. We start by assuming the existence of a training corpus of
N-best lists produced by the recognizer, each list tagged with a
``reference sentence'' that determines which (if any) of the
hypotheses in it was correct. We analyse each hypothesis $H$ in the
corpus using a set of possible KSs, each of which associates some form
of information with $H$.  Information can be of two different kinds.
Some KSs may directly produce a number which can be viewed as a
measure of $H$'s plausibility.  Typical examples are the score which
the recognizer assigned to $H$, and the score for whether or not $H$
received a linguistic analysis (1 or 0 respectively). More commonly,
however, the KS will produce a list of one or more ``linguistic
items'' associated with $H$, for example surface N-grams in $H$ or the
grammar rules occurring in the best linguistic analysis of $H$, if
there was one. A given linguistic item $L$ is associated with a
numerical score through a ``discrimination function'' (one function
for each type of linguistic item), that summarizes the relative
frequencies of occurrence of $L$ in correct and incorrect hypotheses
respectively.  Discrimination functions are discussed in more detail
shortly. The score assigned to $H$ by a KS of this kind will be the
sum of the discrimination scores for all the linguistic items it
finds. Thus each KS will eventually contribute a numerical score,
possibly via a discrimination function derived from an analysis of the
training corpus.

The total score for each hypothesis is a weighted sum of the scores
contributed by the various KSs. The final requirement is to use the
training corpus a second time to compute optimal weights for the
different KSs.  This is an optimization problem which can be
approximately solved using the method described in
\cite{Alshawi+Carter:93}\footnote{A summary can also be found in \cite{SLT}.}.

The most interesting role in the above is played by the discrimination
functions. The intent is that linguistic items which tend to occur
more frequently in correct hypotheses than incorrect ones will get
positive scores; those which occur more frequently in incorrect
hypotheses than correct ones will get negative scores.  To take an
example from the ATIS domain, the trigram {\it a list of} is
frequently misrecognized by DECIPHER as {\it a list the}. Comparing
the different hypotheses for various utterances, we discover that if
we have two distinct hypotheses for the same utterance, one of which
is correct and the other incorrect, and the hypotheses differ by one
of them containing {\it a list of} while the other contains {\it a
list the}, then the hypothesis containing {\it a list of} is nearly
always the correct one. This justifies giving the trigram {\it a list
of} a positive score, and the trigram {\it a list the} a negative one.

We now define formally the discrimination function $d_T$ for a given
type $T$ of linguistic item. We start by defining $d_T$ as a function
on linguistic items. As stated above, it is then extended in a natural
way to a function on hypotheses by defining $d_T(H)$ for a hypothesis
$H$ to be $\sum d_T(L)$, where the sum is over all the linguistic
items $L$ of type $T$ associated with $H$.

$d_T(L)$ for a given linguistic item $L$ is computed as follows. (This
is a slight generalization of the method given in \cite{Collins:93}).
The training corpus is analyzed, and each hypothesis is tagged with
its set of associated linguistic items. We then find all possible
4-tuples $(U, H_1, H_2, L)$ where
\begin{itemize}
\item $U$ is an utterance,
\item $H_1$ and $H_2$ are hypotheses for $U$ exactly one of
which is correct,
\item $L$ is a linguistic item of type $T$ which is associated with exactly
one of $H_1$ and $H_2$.
\end{itemize}
If $L$ occurs in the correct hypothesis of the pair $(H_1, H_2)$, we
call this a ``good'' occurrence of $L$; otherwise, it is a ``bad''
one.  Counting occurrences over the whole set, we let $g$ be the total
number of good occurrences of $L$, and $b$ be the total number of bad
occurrences. The discrimination score of type $T$ for $L$, $d_T(L)$,
is then defined as a function $d(g,b)$. It seems sensible to demand that
$d(g,b)$ has the following properties:
\begin{itemize}
\item $d(g,b) > 0$ if $g > b$
\item $d(g,b) = -d(b,g)$ (and hence $d(g,b) = 0$ if $g = b$).
\item $d(g_1,b) > d(g_2,b)$ if $g_1 > g_2$
\end{itemize}
We have experimented with a number of possible such functions, the
best one appearing to be the following.
\begin{eqnarray*}
d(g,b) = \left\{
                 \begin{array}{ccl}
                  \log_2(2(g + 1)/(g + b + 2))  & if & g < b  \\
                                                0  & if & g = b \\
                  -\log_2(2(b + 1)/(g + b + 2)) & if & g > b
	         \end{array} \right.
\end{eqnarray*}
This formula is a symmetric, logarithmic transform of the function $(g
+ 1)/(g + b + 2)$, which is the expected {\it a posteriori}
probability that a new $(U, H_1, H_2, L)$ 4-tuple will be a good
occurrence, assuming that, prior to the quantities $g$ and $b$ being
known, this probability has a uniform {\it a priori} distribution on
the interval [0,1].

One serious problem with corpus-based measures like discrimination
functions is data sparseness; for this reason, it will often be
advantageous to replace the raw linguistic items $L$ with equivalence
classes of such items, to smooth the data. We will discuss this
further in
Section~\ref{Section:knowledge-sources-used}.

\section{Experiments}
\label{Section:experiments}

This section describes experiments carried out to test
the general methods outlined in the previous section.
Section~\ref{Section:experimental-set-up}
describes the general
experimental set-up, and
Section~\ref{Section:knowledge-sources-used}
the specific knowledge sources used.
Section~\ref{Section:results}
gives the results.

\subsection{Experimental set-up}\label{Section:experimental-set-up}

The experiments were run on the 1001 utterance subset of the ATIS
corpus used for the December 1993 evaluations, which was previously
unseen data for the purposes of the experiments. The corpus,
originally supplied as waveforms, was processed into N-best lists by
the DECIPHER (TM) recognizer. The recognizer used a class bigram
language model. Each N-best hypothesis received a numerical
plausibility score; only the top 10 hypotheses were retained.  The
1-best sentence error rate was about 34\%, the 5-best error rate (i.e.\
the frequency with which the correct hypothesis was not in the top 5)
about 19\%, and the 10-best error rate about 16\%.  Linguistic
processing was performed using a version of the Core Language Engine
customized to the ATIS domain, which was developed under the
SRI-SICS-Telia Research Spoken Language Translator project
\cite{SLT-first-year,SLT,SLT-EUROSPEECH}. The CLE normally assigns a hypothesis
several different possible linguistic analyses, scoring each one with
a plausibility measure. The plausibility measure is highly optimized
\cite{Alshawi+Carter:93}, and for the ATIS domain has an error rate of
about 5\%. Only the most plausible linguistic analysis was used.

The general CLE grammar was specialized to the domain using the
Explanation-Based Learning (EBL) algorithm \cite{EBL-IJCAI-91} and the
resulting grammar parsed using an LR parser \cite{Christer-thesis},
giving a decrease in analysis time, compared to the normal CLE
left-corner parser, of about an order of magnitude. This made it
possible to impose moderately realistic resource limits: linguistic
analysis was allowed a maximum of 12 CPU seconds per hypothesis,
running SICStus Prolog on a Sun SPARC station 10/41.  Analysis that
overran the time-limit was cut off, and corresponding data replaced by
null values.  Approximately 1.2\% of all hypotheses timed out during
linguistic analysis; the average analysis time required per hypothesis
was 2.1 seconds.

Experiments were carried out by first dividing the corpus into five
approximately equal pools, in such a way that sentences from any given
speaker were never assigned to more than one pool\footnote{We would
like to thank Bob Moore for suggesting this idea.}. Each pool was then
in turn held out as test data, and the other four used as training
data. The fact that utterances from the same speaker never occurred
both as test and training data turned out to have an important effect
on the results, and is discussed in more detail later.

\subsection{Knowledge sources used}\label{Section:knowledge-sources-used}

The following knowledge sources were used in the experiments:
\begin{description}
\item[Recognizer score:] The numerical score assigned to each
hypothesis by the DECIPHER (TM) recognizer. This is typically a large
negative integer.
\item[In coverage:] Whether or not the CLE assigned the
hypothesis a linguistic analysis (1 or 0).
\item[Unlikely grammar construction:] 1 if the most plausible
linguistic analysis assigned to the hypothesis by the CLE was
``unlikely'', 0 otherwise. In these experiments, the only analyses
tagged as ``unlikely'' are ones in which the main verb is a form of
{\it be}, and there is a number mismatch between subject and
predicate, e.g.\ {\it ``what is the fares?''}.
\item[Class N-gram discriminants] (four distinct knowledge sources):
Discrimination scores for 1-, 2-, 3- and 4-grams of classes of surface
linguistic items. The class N-grams are extracted after first grouping
some surface words into multi-word phrases, and then replacing some
common words and groups with classes; the dummy words {\it *START*}
and {\it *END*} are also added to the beginning and end of the list
respectively.  Thus, for example, the utterance {\it one way flights to
d f w} would after this phase of processing be {\it *START*
flight\_\-type\_\-adj flights to airport\_\-name *END*}.
\item[Grammar rule discriminants:] Discrimination scores for the grammar rules
used
in the most plausible linguistic analysis of the hypothesis, if there
was one.
\item[Semantic triple discriminants:] Discrimination scores for ``semantic
triples'' in the most plausible linguistic analysis of the hypothesis,
if there was one. A semantic triple is of the form
$(Head_1,Rel,Head_2)$, where $Head_1$ and $Head_2$ are head-words of
phrases, and $Rel$ is a grammatical relation obtaining between them.
Typical values for $Rel$ are ``subject'' or ``object'', when $Head_1$
is a verb and $Head_2$ the head-word of one of its arguments;
alternatively, $Rel$ can be a preposition, if the relation is a PP
modification of an NP or VP. There are also some other possibilities,
cf. \cite{Alshawi+Carter:93}.
\end{description}

The knowledge sources naturally fall into three groups. The first is
the singleton consisting of the ``recognizer score'' KS; the second
contains the four class N-gram discriminant KSs; the third consists of the
remaining ``linguistic'' KSs. The method of \cite{Alshawi+Carter:93}
was used to calculate near-optimal weights for three combinations of
KSs:
\begin{enumerate}
\item Recognizer score + class N-gram discriminant KSs
\item Recognizer score + linguistic KSs
\item All available KSs
\end{enumerate}
In order to facilitate comparison, some other methods were tested as
well. Two variants of the ``highest-in-coverage'' method provided a
lower limit: the ``straight'' method, and one in which the hypotheses
were first rescored using the optimized combination of recognizer
score and N-gram discriminant KSs. This is marked in the tables as
``N-gram/highest-in-coverage'', and is roughly the strategy described
in \cite{BBN:92}. An upper limit was set by a method which selected
the hypothesis in the list with the lower number of insertions,
deletions and substitutions. This is marked as ``lowest WE in 10-best''.

\subsection{Results}\label{Section:results}

Table~\ref{Table:error-rates} shows the sentence error rates for
different preference methods and utterance lengths, using 10-best
lists; Table~\ref{Table:word-error-rates} shows the word error rates
for each method on the full set.  The absolute decrease in the
sentence error rate between 1-best and optimized 10-best with all KSs
is from 33.7\% to 29.9\%, a proportional decrease of 11\%. This is
nearly exactly the same as the improvement measured when the lists
were rescored using a class trigram model, though it should be
stressed that the present experiments used far less training data. The
word error rate decreased from 7.5\% to 6.4\%, a 13\% proportional
decrease. Here, however, the trigram model performed significantly
better, and achieved a reduction of 22\%.

\begin{table}
\begin{center}
\begin{tabular}{|l|c|c|c|c|}
\hline
                            & \multicolumn{4}{|c|}{Max. length (words)} \\
\hline
Preference method           &  8      &  12     &  16    & $\infty$  \\
\hline
\hline
(1-best)                    & 28.3    & 30.4    & 31.9   & 33.7   \\
\hline
highest-in-coverage         & 26.3    & 27.4    & 30.1   & 32.2   \\
\hline
N-gram/highest-in-coverage  & 26.1    & 27.1    & 29.9   & 31.7   \\
\hline
recognizer+N-gram           & 25.3    & 27.8    & 29.7   & 31.6   \\
\hline
recognizer+linguistic KSs   & 23.3    & 24.8    & 27.9   & 30.0   \\
\hline
all available KSs           & 23.5    & 25.4    & 28.1   & 29.9   \\
\hline
(lowest WE in 10-best)      & 12.6    & 13.2    & 14.5   & 15.8   \\
\hline
\hline
(\# utterances in 10-best)  & 442     & 710     & 800    & 843 \\
\hline
(\# utterances)             & 506     & 818     & 936    & 1001 \\
\hline
\end{tabular}
\caption{10-best sentence error rates}
\label{Table:error-rates}
\end{center}
\end{table}

\begin{table}
\begin{center}
\begin{tabular}{|l|c|c|c|c|c|}
\hline
Preference method           & Error rate \\
\hline
\hline
(1-best)                    &  7.4    \\
\hline
highest-in-coverage         &  6.9    \\
\hline
recognizer+N-gram KSs       &  6.8    \\
\hline
N-gram/highest-in-coverage  &  6.7    \\
\hline
recognizer+linguistic KSs   &  6.5    \\
\hline
all available KSs           &  6.4    \\
\hline
(lowest WE in 10-best)      &  3.0    \\
\hline
\end{tabular}
\caption{10-best word error rates}
\label{Table:word-error-rates}
\end{center}
\end{table}

It is apparent that nearly all of the improvement is coming from the
linguistic KSs; the difference between the lines ``recognizer +
linguistic KSs'' and ``all available KSs'' is not significant.  Closer
inspection of the results also shows that the improvement, when
evaluated in the context of the spoken language translation task, is
rather greater than Table~\ref{Table:error-rates} would appear to
indicate. Since the linguistic KSs only look at the abstract semantic
analyses of the hypotheses, they often tend to pick harmless syntactic
variants of the reference sentence; for example {\it all of the} can
be substituted for {\it all the} or {\it what are ...} for {\it which
are ...}. When syntactic variants of this kind are scored as correct,
the figures are as shown in Table~\ref{Table:error-rates-acceptable}.
The improvement in sentence error rate on this method of evaluation is
from 28.8\% to 22.8\%, a proportional decrease of 21\%. On either type
of evaluation, the difference between ``all available KSs'' and any
other method except ``recognizer + linguistic KSs'' is significant at
the 5\% level according to the McNemar sign test \cite{Powell:76}).

\begin{table}
\begin{center}
\begin{tabular}{|l|c|c|c|c|}
\hline
                            & \multicolumn{4}{|c|}{Max. length (words)} \\
\hline
Preference method           &  8      &  12     &  16    & $\infty$  \\
\hline
\hline
(1-best)                    & 24.3    & 26.0    & 27.5   & 28.8   \\
\hline
highest-in-coverage         & 20.4    & 21.5    & 23.7   & 25.3   \\
\hline
recognizer+N-gram KSs       & 20.4    & 22.5    & 23.8   & 25.2   \\
\hline
N-gram/highest-in-coverage  & 19.0    & 20.5    & 22.6   & 24.1   \\
\hline
recognizer+linguistic KSs   & 17.6    & 19.6    & 21.7   & 23.5   \\
\hline
all available KSs           & 17.6    & 19.6    & 21.5   & 22.8   \\
\hline
(lowest WE in 10-best)      & 11.3    & 12.0    & 13.0   & 14.0   \\
\hline
(\# utterances)             & 506     & 818     & 936    & 1001 \\
\hline
\end{tabular}
\caption{10-best sentence error rates
counting acceptable variants as successes}
\label{Table:error-rates-acceptable}
\end{center}
\end{table}

One point of experimental method is interesting enough to be worth a
diversion. In earlier experiments, reported in the notebook version of
the paper, we had not separated the data in such a way as to ensure
that the speakers of the utterances in the test and training data were
always disjoint. This led to results which were both better and also
qualitatively different; the N-gram KSs made a much larger
contribution, and appeared to dominate the linguistic KSs. This
presumably shows that there are strong surface uniformities between
utterances from at least some of the speakers, which the N-gram KSs
can capture more easily than the linguistic ones. It is possible that
the effect is an artifact of the data-collection methods, and is
wholly or partially caused by users who repeat queries after system
misrecognitions.

There were a total of 88 utterances for which there was some
acceptable 10-best hypothesis, but where the hypothesis chosen by the
method which made use of all available KSs was unacceptable. In order
to get a more detailed picture of where the preference methods might
be improved, these utterances were inspected and categorized into
different apparent causes of failure. Four main classes of failures
were considered:
\begin{description}
\item[Apparently impossible:] There is no apparent
reason to prefer the correct hypothesis to the one chosen without
access to intersentential context or prosody. There were two main
subclasses: either some important content word was substituted by
an equally plausible alternative (e.g.\ ``Minneapolis'' instead of
``Indianapolis''), or the utterance was so badly malformed that
none of the alternatives seemed plausible.
\item[Coverage problem:] The correct hypothesis was not in implemented
linguistic coverage, but would probably have been chosen if it had
been; alternately, the selected hypothesis was incorrectly classed
as being in linguistic coverage, but would probably not have been
chosen if it had been correctly classified as ungrammatical.
\item[Clear preference failure:] The information needed to make the
correct choice appeared intuitively to be present, but had not been
exploited.
\item[Uncertain:] Other cases.
\end{description}
The results are summarized in Table~\ref{Table:failure-causes}.

\begin{table}
\begin{center}
\begin{tabular}{|l|r|}
\hline
Apparently impossible                       &  14     \\
\hline
Coverage problems                           &  44     \\
\hline
Clear preference failure                    &  21     \\
\hline
Uncertain                                   &  9     \\
\hline
\end{tabular}
\caption{Causes of N-best preference failure}\label{Table:failure-causes}
\end{center}
\end{table}
At present, the best preference method is in effect able to identify
about 40\% of the acceptable new hypotheses produced when going from
1-best to 10-best. (In contrast, the ``highest-in-coverage'' method
finds only about 20\%).  It appears that addressing the problems
responsible for the last three failure categories could potentially
improve the proportion to something between 70\% and 90\%.  Of this
increase, about two-thirds could probably be achieved by suitable
improvements to linguistic coverage, and the rest by other means. It
seems plausible that a fairly substantial proportion of the failures
not due to coverage problems can be ascribed to the very small
quantity of training data used.

\section{Summary and conclusions}
\label{Section:summary}

We have described a simple and uniform architecture for combining
different knowledge sources to create an N-best preference method.
The method can easily absorb new knowledge sources as they become
available, and can be automatically trained. It is economical with
regard to training material, since it makes use of both correct and
incorrect recognizer hypotheses. It is in fact to be noted that over
80\% of the discrimination scores are negative, deriving from
incorrect hypotheses. The apparent success of the method can perhaps
most simply be explained by the fact that it attempts directly to
model characteristic mistakes made by the recognizer. These are often
idiosyncratic to a particular recognizer (or even to a particular
version of a recognizer), and will not necessarily be easy to detect
using more standard language models based on information derived from
correct utterances only.

We find the initial results described here encouraging, and in
the next few months intend to extend them by training on larger
amounts of data, refining existing knowledge sources, and adding new
ones.  In particular, we plan to investigate the possibility of
improving the linguistic KSs by using partial linguistic analyses when
a full analysis is not available. We are also experimenting with
applying our methods to N-best lists which have first been rescored
using normal class trigram models. Preliminary results indicate a
proportional decrease of about 7\% in the sentence error rate when
syntactic variants of the reference sentence are counted as correct;
this is significant according to the McNemar test. Only the linguistic
KSs appear to contribute. We hope to be able to report these results
in more detail at a later date.

\end{document}